\documentclass{article}
\usepackage{arxiv}
\usepackage[utf8]{inputenc}
\usepackage[T1]{fontenc}  
\usepackage{hyperref}    
\usepackage{url}      
\usepackage{graphicx}
\usepackage{doi}
\usepackage[utf8]{inputenc}
 
\usepackage[backend=biber,style=apa,url=true]{biblatex}
\usepackage{hyperref} 
\title{Mediating Modes of Thought: LLM's for design scripting.}

\author{ 
  \href{https://orcid.org/0009-0006-0647-1411}{\includegraphics[scale=0.06]{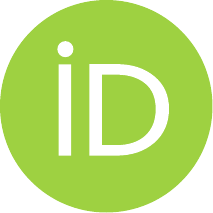}\hspace{1mm}Moritz~Rietschel} \\
	Department of Architecture\\
	University of California, Berkeley\\
	\href{mailto:rietschel@berkeley.edu}{\texttt{rietschel@berkeley.edu}} \\
	\And
	Fang~Guo \\
	Department of Architecture\\
	University of California, Berkeley\\
	\href{mailto:fang_guo@berkeley.edu}{\texttt{fang\_guo@berkeley.edu}} \\
	\AND
 	\href{https://orcid.org/0000-0003-2631-676X}{\includegraphics[scale=0.06]{orcid.pdf}\hspace{1mm}Kyle~Steinfeld} \\
	Department of Architecture\\
	University of California, Berkeley\\
	\href{mailto:ksteinfe@berkeley.edu}{\texttt{ksteinfe@berkeley.edu}} \\
}

\renewcommand{\shorttitle}{\textit{Mediating Modes of Thought}}

\hypersetup{
pdftitle={Mediating Modes of Thought},
pdfsubject={q-bio.NC, q-bio.QM},
pdfauthor={Moritz.~Rietschel, Fang~Guo, Kyle~Steinfeld},
pdfkeywords={Design Thinking and Applied Research, Design Cognition, Human-AI Interaction, Generative AI for Design, Computational Creativity, Human-Centered AI},}

\begin{document}
\maketitle

\begin{abstract}
	Architects adopt visual scripting and parametric design tools to explore more expansive design spaces \parencite{coates_programmingarchitecture_2010}, refine their thinking about the geometric logic of their design \parencite{woodbury_elements_2010}, and overcome conventional software limitations \parencite{burry_scripting_2011}. Despite two decades of effort to make design scripting more accessible, a disconnect between a designer's free ways of thinking and the rigidity of algorithms remains \parencite{burry_scripting_2011}. Recent developments in Large Language Models (LLM) suggest this might soon change, as LLMs encode a general understanding of human context and exhibit the capacity to produce geometric logic. This project speculates that if LLMs can effectively mediate between
user intent and algorithms they become a powerful tool to make scripting in design
more widespread and fun. We explore if such systems can interpret natural language prompts to assemble geometric operations relevant to computational design scripting.
In the system, multiple layers of LLM agents are configured with specific context to infer the user intent and construct a sequential logic. Given a user’s high-level text prompt, a geometric description is created, distilled into a sequence of logic operations, and mapped to software-specific commands. The completed script is constructed in the user's visual programming interface. The system succeeds in generating complete visual scripts up to a certain complexity but fails beyond this complexity threshold. It shows how LLMs can make design scripting much more aligned with human creativity and thought. Future research should explore conversational interactions, expand to multimodal inputs and outputs, and assess the performance of these tools.
\end{abstract}

% keywords can be removed

\section{Introduction}
This paper proposes Large Language Models (LLMs) as mediators in computational design processes, suggesting them as effective bridges between human intuition and algorithmic logic in design. It hopes to uncover new paths in the ongoing effort to enable designers to wield the powerful and inherently algorithmic tools of the digital age.

Parametric modeling is a driver of the complex geometric style of contemporary architecture since it enables the exploration of design spaces that are too complex to manage or explore manually. Starting around the turn of the century, end user programming in design disciplines marked a clear departure from the previous Computer Aided Design tools, as it moved beyond the predefined operations created by software engineers, and turned designers into digital tool makers \parencite{burry_scripting_2011}. The results are algorithms crafted for specific design problems, and usually not re-employed to new problems without adapting, editing and recombining them. The rise of this “scripting culture” represents one of the defining features of the past few decades in architectural practice, and holds huge potential for an expansion of design exploration. However, it remains limited in distribution among designers because it is difficult to master and algorithms do not allow for ambiguity that is common at the early design stages.

Designers and design students frequently express frustration with the steep learning curve of parametric design tools and algorithmic logic. There is a fundamental difference in these tools from the familiar CAD suites in that they require abstraction and conceptualization, where there previously was immediate interaction and modification \parencite{burry_scripting_2011}. Traditional digital models
are typically built up and edited through graphical user interfaces, where users combine interaction in the 3D viewport with a selection of commands from the graphical user interface (GUI). In many popular systems, these operations are not recorded, leading to irreversible model changes. CAD software is always based on mathematical operations and digital representations of models, but a designer’s interaction with them mimics the immediacy of real world interactions of drawing and modeling.

With scripting the designer no longer designs the model itself, but instead the chain of operations and relationships that lead from specified inputs to a design space exploration. This requires a shift to an algorithmic mode of thought. A lot of confusion and frustration stems from this unfamiliar and radically different mode of design, in which the design algorithm relies on complete, error free sets
of operations to produce sensible, visual output to a user.

The trade off is especially steep in the first stages of design, where sketching algorithms takes far too much time and metal load, and the deterministic structure required for algorithmic logic is impossible to define because of the very nature of this design phase.

Design tools that employ machine learning (ML) could change this dynamic. Because such systems learn from experience and respond to new contexts by inferring from observed patterns, we project that such tools may be better aligned with a designer’s mode of thinking in comparison
to traditional CAD and parametric modeling tools \parencite{steinfeld_dreams_2017}. Large language models offer a particular opportunity in the context of parametric modeling. LLMs have internalized representations of human context and can also produce code and logical steps. They present a promising new avenue for mediating the fundamental difference in thinking, potentially transforming how designers interact with algorithmic tools.

This research explores this new domain of interaction, working towards aligning parametric workflows with a designers mode of thinking. We present a prototype system to test this idea that allows the user to create design scripts from natural language prompts, demonstrating the potential of LLMs in generating and reasoning about design scripts. The results from our system study show the promising ability of language models to infer design intent and construct a geometric logic from it. Our prototype system allows a new and different interaction with design scripts. It points to a future of algorithmic design mediators.

\section{State of the Art}
\label{sec:Stateoftheart}
Our proposed approach to generative design is related to efforts to make coding accessible, to introducing generative AI in computational design, and the theory of design thinking and how it is reflected in computational design tools. The following outlines the current state of these fields.

\subsection{Visual Scripting}
In visual programming interfaces, flow charts are assembled from a fixed set of components, illustrating the logical flow of the program. They have a rich history as a tool to make programming more accessible, often employed as an introduction to programming logic like the website Scratch \parencite{maloney_scratch_2004}. However, while graphical user interfaces have become a routine interface for computing, they are a far abstraction from the underlying algorithmic processes \parencite{woodbury_elements_2010}. This disconnect created a need for end user programming tools in domain specific software as diverse as MS Office, visual arts \parencite{vvvv_software}, music \parencite{max_software}, and architecture \parencite{rutten_grasshopper3d_2007} to re-expose the user to powerful computational logic through a visual interface.

\subsection{Generative AI in tools}
Generative artificial intelligence (GAI) has made significant advantages in recent years, and has been explored as a tool within or in addition to existing software. A widespread application for LLMs in particular are coding assistants for computer programming. These built in helpers such as GitHub CoPilot, Amazon Q and IBM’s Watson are trained from massive collections of text \parencite{radford2019language} to generate custom logic, fix errors and explain code. Generative AI tools are fundamentally different from past computational tools because their output is non-deterministic, creating unpredictability and imprecision. Applying them to design is especially interesting because they draw from experience to generate new solutions, can handle imprecisions in input data and generate output that goes beyond predetermined operations. With the rising interest in generative AI, such systems are being explored in computational design, often as interventions during the design process or as tools for modeling. For example, in the design process, the integration of generative AI has been shown to be useful in providing inspiration and help explore a design space \parencite{yousif_towards_2022}. In design for physical making, such systems have been tested to produce design specification, design models, assembly logic's and manufacturing specifications \parencite{makatura_how_2023}. The models have also been used as an educational tool in computational design, to instruct a user, help grasp concepts and assess user understanding through the power of conversational interfaces \parencite{milrad_ai4architect_2023}. Procedural and parametric design workflows in particular exhibit complementary characteristics to learning based tools, offering interpretability, manipulation and composition capabilities \parencite{ritchie_neurosymbolic_2023}.

\subsection{Immediate Precursor to the Project}
In architecture, the Grasshopper environment is a very successful example for visual programming that has seen widespread adoption, and itself has become a platform, supporting an extensive range of plugins for domain specific scripting. We built our system on top of the existing open source Grasshopper plugin GHPT which generates scripts in a similar approach to ours. This project created
a JSON schema that defines the script components and their connections, and the code that parses them onto the canvas. The plugin performs a text completion API call to an LLM that includes instructions, example scripts in the JSON schema, and the users prompt and then tasks the LLM to complete a new JSON schema based on it. Their project does not employ task separation, step-by-step thinking or multiple LLM calls. Still, the plugin manages to generate simple scripts and snippets in Grasshopper code. Our prototyped system was built on top of their open source code base, expanding it to increase the robustness of the JSON parsing, but our reasoning approach is developed separately from theirs \parencite{sykes_enmerk4r_2023}.

\begin{figure}
	\centering
	\includegraphics[width=0.5\textwidth]{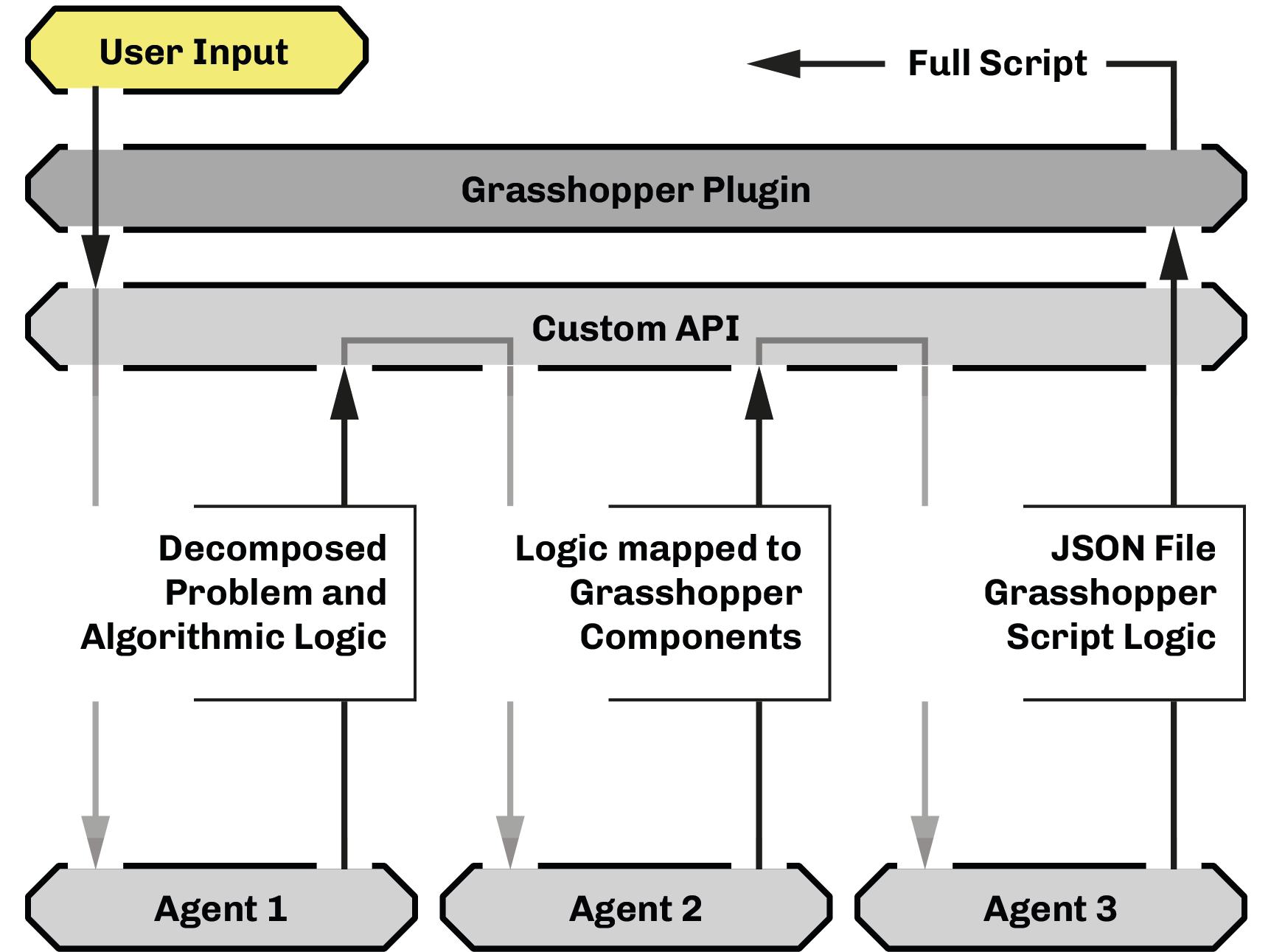}
	\caption{A diagram that outlines the flow of the prototyped system, leading from the user input to the final script.}
	\label{fig:fig1}
\end{figure}

\section{System}
\label{sec:system}
We propose systems in which LLMs support a designer by interpreting their intent from natural language descriptions and corresponding design logic. Our prototype system tests the capabilities of LLMs to serve as such interpreters and logic constructors. It infers the design intent from a text input, decides on inputs to parameterise, and constructs the relevant geometric logic script. This algorithm is then mapped to software specific components of Grasshopper and produced directly within the user's GUI. We built this tool using the LLM GPT4o from OpenAI.

\subsection{LLM Techniques}
Since large language models are inherently unreliable and difficult to control, precise prompting is essential to this prototype’s functionality, playing a key role in achieving our results and reproducibility. The LLM we employ (GPT-4o) is a general model trained on vast amounts of unspecific data, allowing it to exhibit an understanding of human context but making it difficult to confine it to a domain. We provide precise descriptions of the functionalities of computer aided design tools, Euclidean geometry, and the tasks for which architects employ parametric design as context for the models' output generation. Specifically, we instructed the system to produce abstract representations of the user prompt within the space and context of CAD tools, disregarding considerations outside of geometry and algorithmic logic. Giving precise context and instructions is essential for our approach to using LLMs. Instructing the model to output text detailing step-by-step chains of thought improves its ability to reason and creates a more methodical mode of thinking and inferring for the model \parencite{li_chain_2024}. We also instruct our model to take mental notes throughout the process, to force reasoning steps during generation \parencite{anthropic_let_2024}. We split the task into three separate LLM interactions to ensure only immediately relevant context in the LLM’s context window \parencite{wu_stateflow_2024} and order our instructions according to the intended approach \parencite{chen_premise_2024}.

\subsection{System Architecture}
The system architecture includes three key layers: the LLM agents, the API handling message exchanges, and the Grasshopper plugin parsing the generated JSON to the canvas. We built our user facing system by expanding on the open source project GHPT \parencite{sykes_enmerk4r_2023}. The user provides a text prompt, which is sent via the API to the first LLM agent to deduce the user's design intent and define inputs and the geometric logic for a script. These operations are mapped to a Grasshopper specific script, and translated to a JSON file by the subsequent agents. The JSON is returned to the Grasshopper environment to produce the script.

\subsection{Agent Functionalities}
The first agent understands the user intent from the prompt, instructed with the specific context mentioned above. After describing the inferred design intent, the system moves on to assume practical inputs to parameterise; values within the design goal that a user might want to edit in the specific context of design space exploration, and assigns values and ranges for these values. At this point, the system has a goal and specified inputs; a well defined context to construct the geometric logic. It then constructs a chain of geometrical operations that lead from the specified inputs to the goal geometry. The complete output of this first agent is then a set of inputs, operations and a description of the goal geometry.

The second agent maps the created logic from the firstto a set of specified components or operations from the Grasshopper environment. It also adjusts the script to this environment’s specific, flow-based constraints, breaking down any loops or repetitions in the logic. It outputs a chain of operations that includes each component the connections between them; the final logic for the script.

The third LLM agent takes the component chain produced by the second agent and converts it to a specified schema in the JSON language. It must follow the specified structure precisely for successful parsing by the plugin later. The schema separately lists every component and every connection. This schema was adopted from the GHPT project. The LLM we used for this agent, GPT-4o by OpenAI, allows us to ensure that the model’s response is in valid JSON format, but does not guarantee a specific schema.

\subsection{Instantiating Components}
Our system relies on the open source project GHPT that implemented a Grasshopper Plugin to parse the JSON schema to a script on the user's canvas. It uses the GrasshopperSDK to fetch each component's information by name, places it on the canvas and then assembles the connections to complete the script. We implemented changes in the logic of fetching components and adding connections, to increase the robustness of fetching the correct component and connection. We also limited the possible list of components to the ones listed in the agent's prompts, to reduce complexity and make the system more reliable.

\section{Results}
\label{sec:results}
The following section provides three examples where we recorded the output of the first and second agent, as well as a screenshot of the final script in the user interface of Grasshopper. We chose to test the system with high level prompts to probe the LLMs internal understanding of the world and its geometries, broadly assessing the systems utility in an architecture context. Full responses and the produced JSON files can be found on our website.

\begin{figure}
	\centering
	\includegraphics[width=0.8\textwidth]{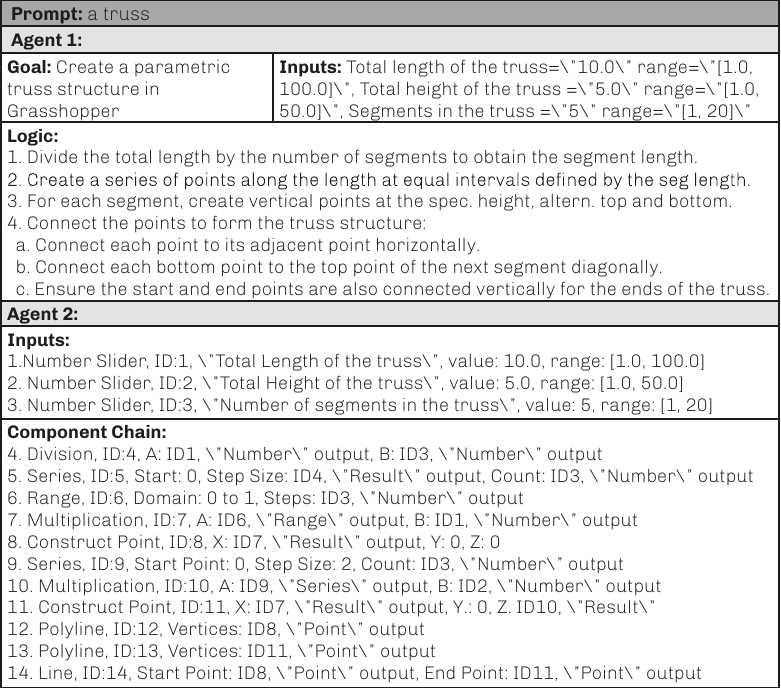}
	\caption{A table of shortened responses from the first two agents to the prompt “a truss”. The result of the first agent serves as the input for the second.}
	\label{fig:fig2}
\end{figure}

\begin{figure}
	\centering
	\includegraphics[width=0.8\textwidth]{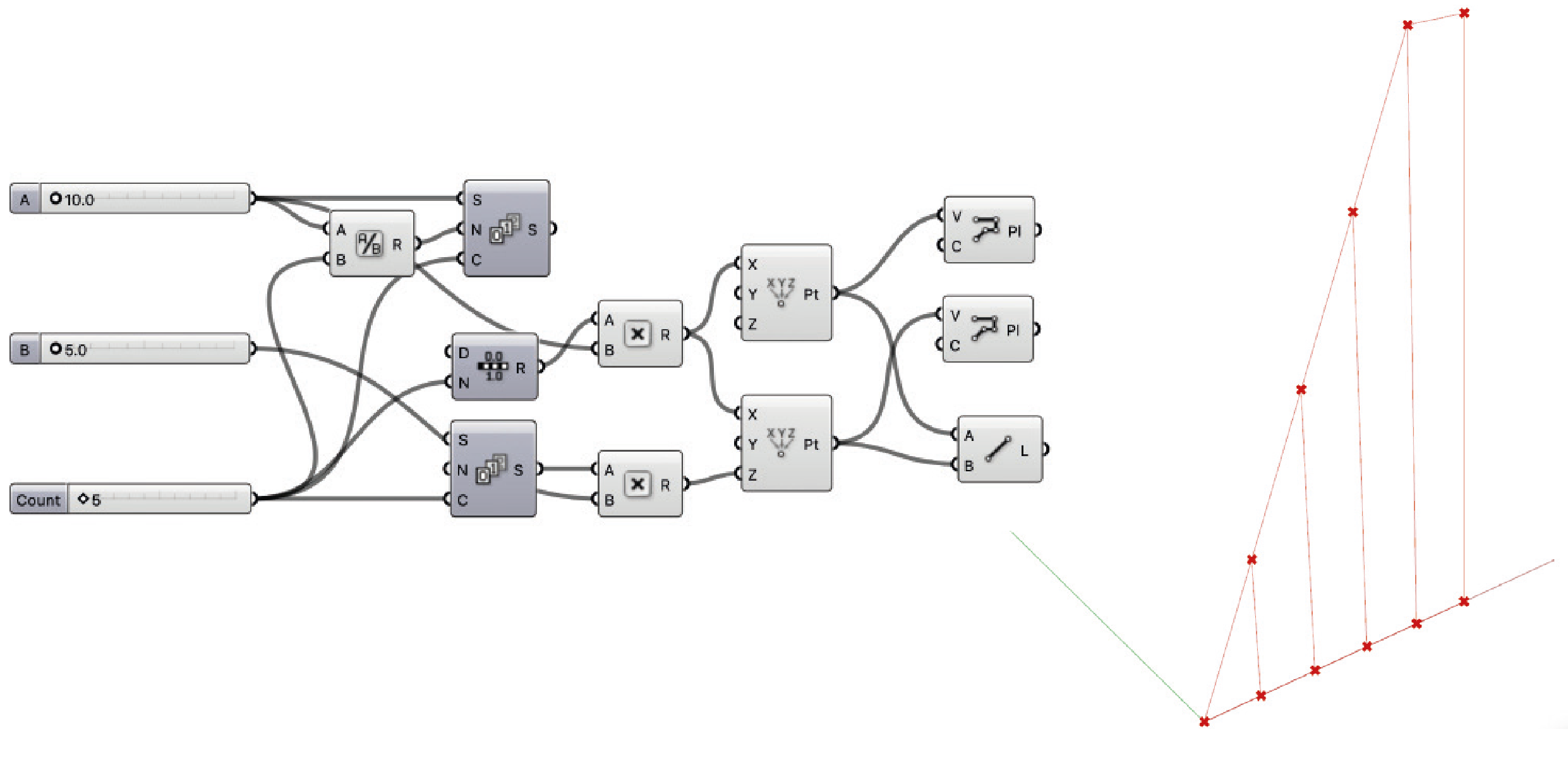}
	\caption{A visual script and CAD preview that was parsed from the third agent's JSON onto the Grasshopper canvas for the prompt “a truss”.}
	\label{fig:fig3}
\end{figure}

\subsection{Result 1: Truss}
In this case study \ref{fig:fig2}, the system was instructed to produce a parametric script for a truss. Without any additional context, a truss could refer to several geometries, and the LLM must assume what geometry to create. From the textual logic it becomes clear that the intended geometry is a straight truss with diagonal connections. The inputs selected for this straight truss are the length and height dimensions and the number of cross-connections. 

The LLM constructs a set of points relative to the length of the truss. There is no mention of the start and end points or constructing the lines that make up the top and bottom edges of the truss. Instead, the logic describes how to produce the diagonal cross-connections.

The second agent maps this logic to Grasshopper by producing several lists of points and connecting them with polylines. The logic successfully produces two lists of points, the top and bottom lines of the truss and the cross-connections. However, the cross-connections are not diagonal as described in the logic, and the truss is not straight but an inclined, half-gable truss.

It performs the division described in the logic and then creates a series of points but never feeds into the script’s output, which makes it a lost node. Even though the script does not match the first agent’s intention of a straight truss, the parametric logic is intact, and we can imagine this script serving as a starting point for a designer’s exploration of trusses.

\subsection{Result 2: Umbrella}
\begin{figure}
	\centering
	\includegraphics[width=0.8\textwidth]{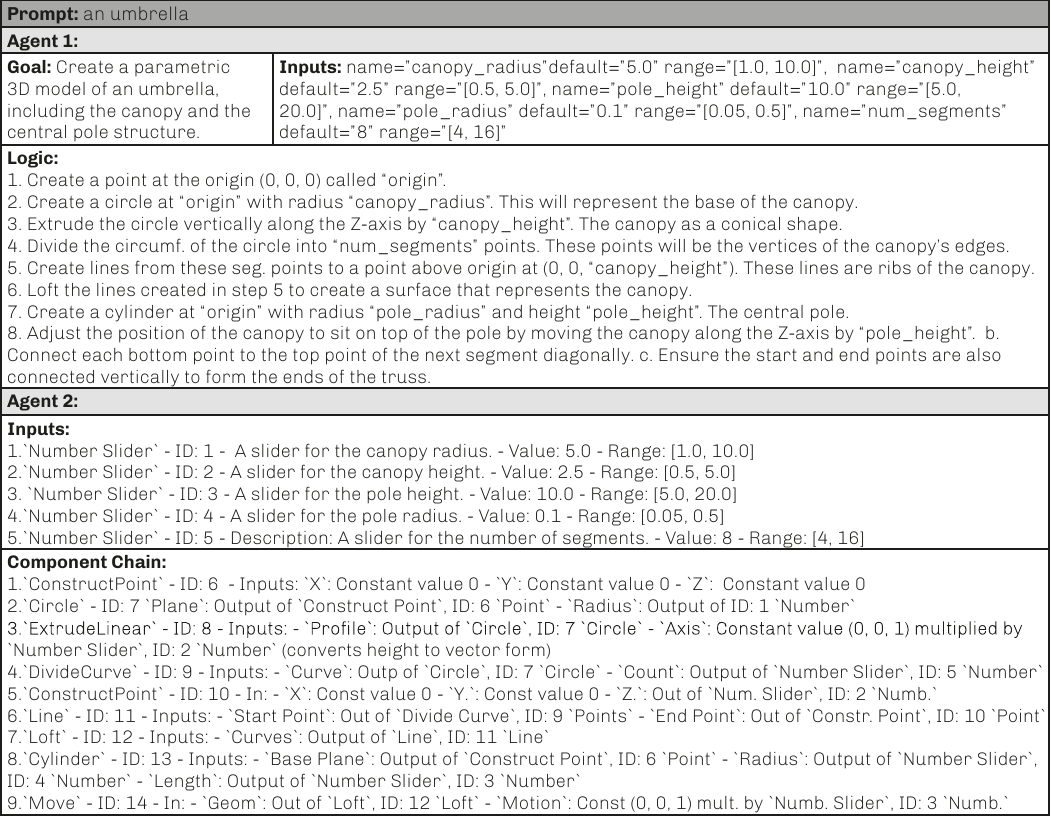}
	\caption{A table of shortened responses from the first two agents to the prompt “an umbrella”. The result of the first agent serves as the input for the second.}
	\label{fig:fig4}
\end{figure}

For the second case study \ref{fig:fig4}, we asked the system to produce an umbrella. This prompt requires the system to surface the geometric relationship of a structure that we intuitively associate with an umbrella. It produces the canopy and the central pole and defines the necessary inputs.

This logic proceeds in a classic CAD workflow, beginning by constructing an origin point and drawing the geometries from there. It includes a false extrusion logic, but the second agent ignores that. Instead, the lofting of canopy segments produces the correct surface, and the central pole is added. It is unclear why the LLM added two different approaches to produce the canopy surface.

The second agent maps the number slider inputs, the constructed point, and other commands. It also attempts to reproduce the false circle extrusion but fails in the translation of the logic. The ‘ExtrudeLinear’ component fails because it requires an axis input instead of a number.
The other error produced by the agent is the failed ‘Move’ component at the end of the logic, which would have required a vector instead of a number input. Even though the second agent’s instructions include a list of components and their required inputs, the input format matching errors are common.

The umbrella logic includes two failed components yet successfully produces a canopy with a dynamic number of segments, a relatively complex logic. Users with some familiarity with Grasshopper could fix the errors quickly and use the umbrella script in their work.

\begin{figure}
	\centering
	\includegraphics[width=0.8\textwidth]{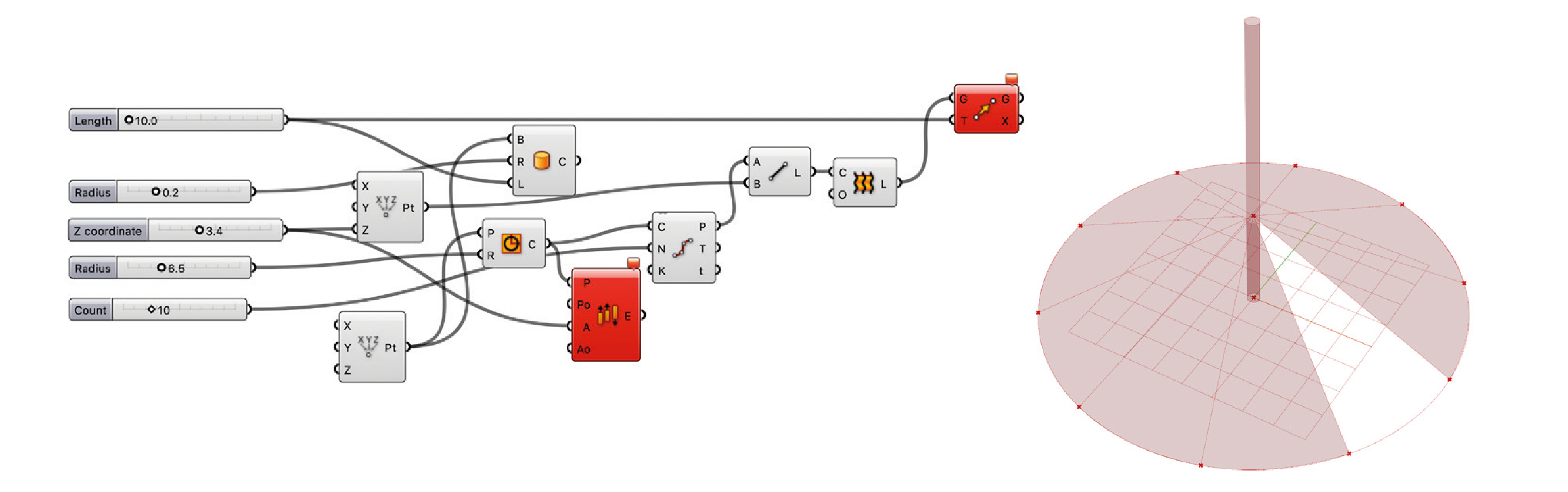}
	\caption{A visual script and CAD preview that was parsed from the third agent's JSON onto the Grasshopper canvas for the prompt “an umbrella”.}
	\label{fig:fig5}
\end{figure}

\subsection{Result 3: Suspension Bridge}
\begin{figure}
	\centering
	\includegraphics[width=0.8\textwidth]{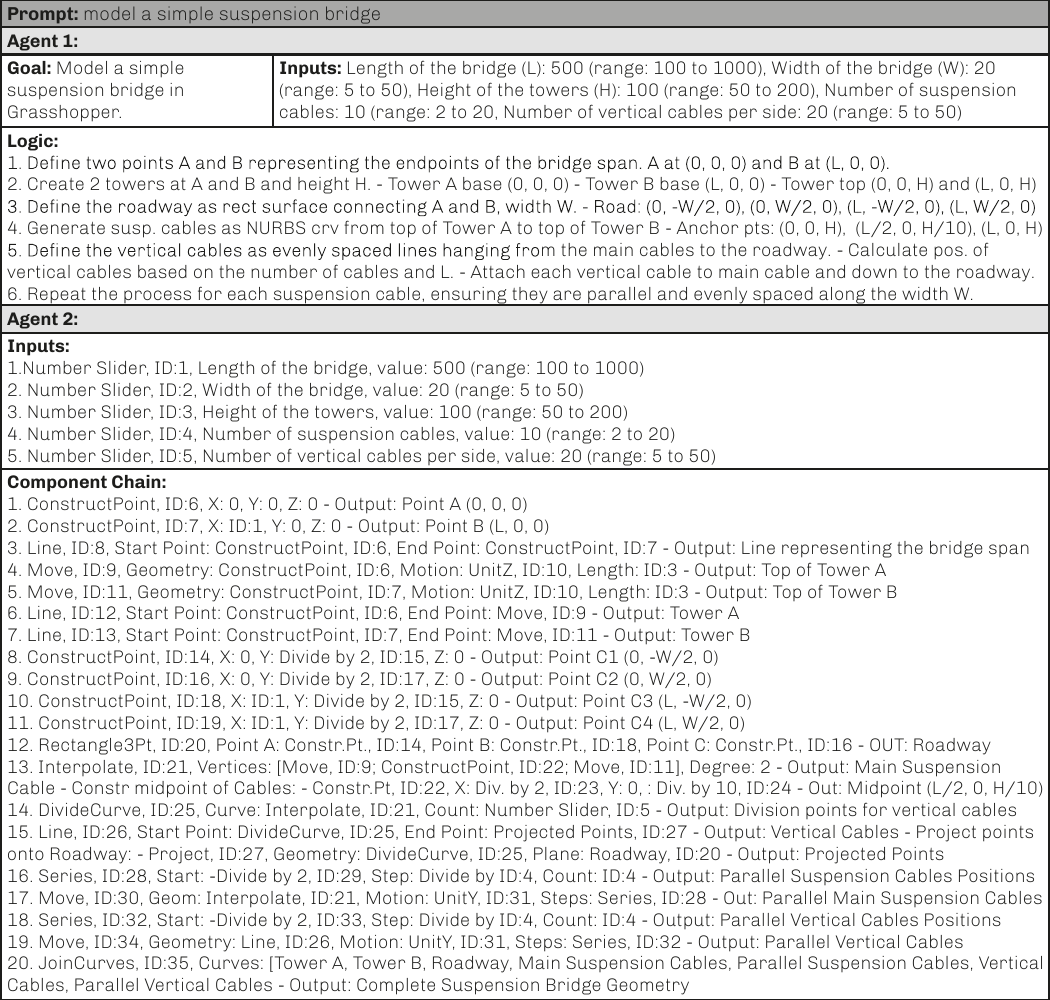}
	\caption{A table of shortened responses from the first two agents to the prompt “a simple suspension bridge”. The result of the first agent serves as the input for the second.}
	\label{fig:fig6}
\end{figure}

The final case study \ref{fig:fig6} requests a suspension bridge, a much more complex geometry than the previous two results. Reproducing such a geometry requires detailed inferences of what the user is looking for by a suspension bridge, and this model decided to produce a suspension bridge with vertical towers.

The logic begins by constructing towers at the start and end of the bridge and the roadway between them. It then defines the suspending cables as hanging from the towers to anchor points in the middle of the bridge length and vertical cables from it to the roadway. Since the inputs include a parameter for a number of suspension cables, the logic adds them in parallel and evenly spaced over the bridge.

When mapping the logic to Grasshopper, all move operations fail because their inputs are incorrectly defined.Instead of constructing towers, this agent attempts to construct lines, simplifying the logic. The suspension cable logic succeeds in producing a NURBS curve, but the anchor point is located directly above the endpoint instead of the middle of the span.

Overall, modeling a suspension bridge proved too complex for the system, resulting in a strange and confusing script and a visual output that bears little resemblance to a bridge. The many missing connections and sink nodes show that the complexity of a suspension bridge is too high for the current system.

\begin{figure}
	\centering
	\includegraphics[width=0.8\textwidth]{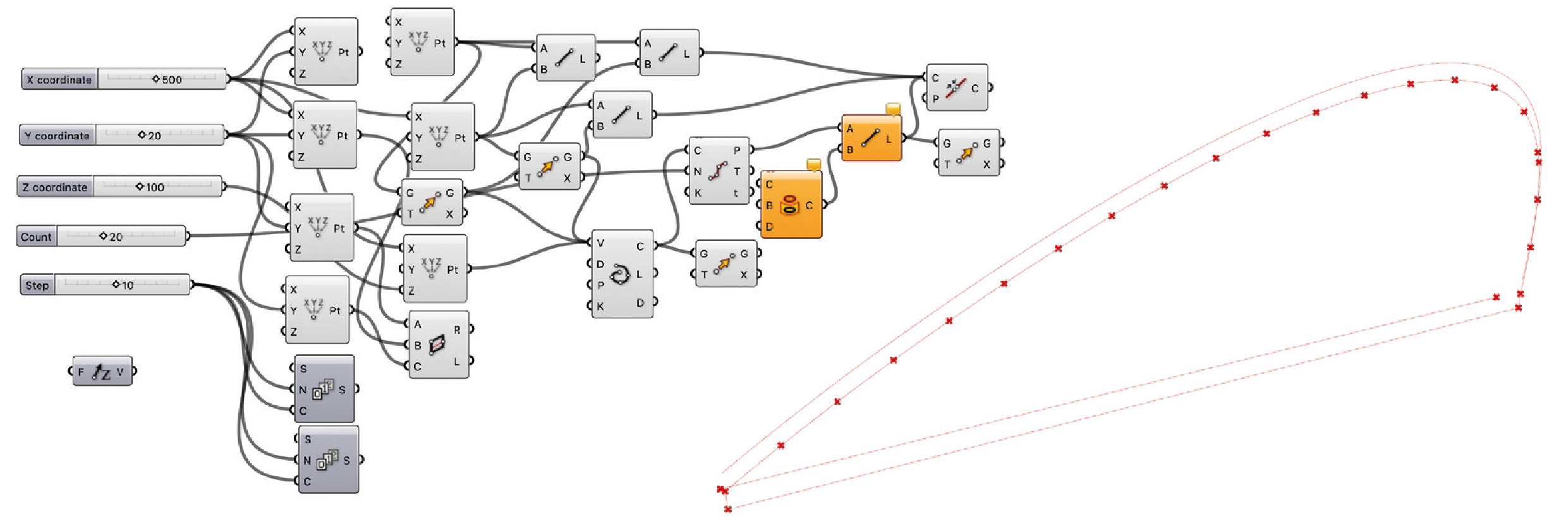}
	\caption{A visual script and CAD preview that was parsed from the third agent's JSON onto the Grasshopper canvas for the prompt “a simple suspension bridge”.}
	\label{fig:fig7}
\end{figure}

\subsection{Complete System: End-to-End Performance}
The complete system integrates all agents and the instantiating process to generate a parametric script
from a natural language prompt. The results of this whole system are mixed. Errors accumulate in the system in every step and become more likely in longer, more complex scripts.

The first agent displayed an impressive learned understanding of geometries based on short, minimal prompts. It also outlines a useful logical sequence to construct them despite large variations in their precision. The second agent is prone to making errors where
the Grasshopper environment differs from standard programming logic, such as its characteristic forward propagation, lack of loops, and missing global variables. Despite being instructed to pick only from the provided list of components, this agent can hallucinate non-existent components, leading to errors. The second agent was observed to be most successful with straightforward logic and becomes worse as the script becomes longer and more complex. Translating the logic to JSON often produces errors. Specifically, the language model struggles with maintaining consistency and completeness in the JSON format. Common issues are missing connections between components, wrong parameter or component names, and deviation from the schema. 

Errors add up and reduce the likelihood of success, usually leading to a few errors even in successful scripts. It becomes clear that the system is better at coming up with logic than mapping its logic to the specific domain of Grasshopper, the weakest element being the JSON generation, since this relies on exact names and structures. Notable limitations of the current system are incorrect component selection due to their ambiguous natural language names and missing connections in complex scripts. The final Grasshopper script often includes a few wrong connections and components. Despite these challenges, the system consistently produced at least parts of a logic structure in the output.

\section{Discussion}
\label{sec:discussion}
The prototype system can infer designer intent from prompts and produce computational logic from it. It showcases how LLMs can mediate between human designers and their computational tools, translating high- level prompts into algorithmic logic. We project that this capability can make parametric design more accessible and easy. However, limitations in handling complex scripts and maintaining consistency indicate a need for further refinement, limiting its current practical utility.

It becomes clear that the LLM can draw from its learned data to recreate reasonably complex geometric relationships in a parametric, ready-to-use script. This can become a useful functionality for a parametric designer, serving as a customized library for logic sequences. The LLM can recreate abstract patterns of logic encountered in its vast data, making a potentially unlimited database of such patterns available within a design environment. In the future, an integrated LLM might explain the logic of a script to the user, answering questions in a conversational interface and easing the steep technical learning curve of scripting logic. This approach can be used to implement LLMs into any visual programming environment where the generated logical structure will be easy to grasp, edit, and extend and it will conform to predefined components. While our approach does not extend the scope of the tools, it can significantly lower the barriers of entry and spread them to designers currently reluctant to employ design scripting.

We acknowledge that LLMs are inherently unreliable and that their reasoning capability is questionable. They are still unable to self correct their reasoning \parencite{huang_large_2024}. While this limitation does not pose a problem when recreating well-known geometric logic, if the LLM fails to reason independently it limits the complexity of logic that can be constructed. Additionally, we project that offloading algorithm creation to a system such as ours may reduce originality and understanding in the design process, and expect that our system is unlikely to construct novel, unique, and coherent geometric algorithms as an experienced parametric designer might. 

Future research should focus on improving the reliability and quality of the generated output, exploring conversational user interfaces, and expanding to multi- modal inputs. We imagine a system integrated deeply within the design environment, with an awareness of all relevant contexts. Pragmatic improvements to our code are enhancing prompting techniques, error correction, retrieving augmented generation for component info, and making the parsing more robust to faulty JSON. Another crucial step is conducting user studies and collecting feedback to assess the system's utility for designers.

\section{Conclusion}
\label{sec:conclusion}
This project shows the potential of LLMs to transform the parametric design process by combining their learned understanding of the world with the ability to write code. Our prototype showed a fundamentally different construction of algorithms in design, moving towards design tools that are better aligned with how designers think. Our prototype produced parametric scripts or snippets for a truss, an umbrella, and a suspension bridge from minimal textual inputs. While the prototyped system shows promise, further development is needed to handle complex design tasks reliably, which is where we expect the tool to be most valuable.

Our project is part of a larger push in computational design from deterministic tools towards tools that think more like designers themselves. We had set out to explore if
the learned understanding of human context within large language models made them helpful mediators between human designers and design scripts. Our results indicate that these models can produce a geometric understanding of a high-level user input and produce a sound parametric logic, showing effective translation between the two. This research suggests a promising future for LLMs as mediators in design, leveraging their knowledge of our world and language to support our use of computational tools. These new tools will make programming in design as accessible and broadly employed as the current non-parametric CAD tools.

\section{Acknowledgements}
\label{sec:acknowledgements}
We would like to thank Mike Ren and Ada Fang for their early contributions to the project. We would also like to thank ZeroWidth for providing us with free access to LLM’s, and Prof. Niloufar Salehi from the University of California Berkeley’s School of Information for her guidance during the initial writing of this paper. This paper has been accepted for presentation at the ACADIA 2024 conference.

\section{Additional Resources}
\label{sec:additionalresources}
Full prompts, results and code on morietschel.github.io/MMOT

\printbibliography

\end{document}